\documentclass{article} 
\usepackage{iclr2025_conference,times}

\usepackage{rotating}
\usepackage{booktabs}
\usepackage{float}

\usepackage{tikz}
\usetikzlibrary{positioning, calc, shapes.geometric, decorations.pathreplacing}

\usepackage{amsmath,amsfonts,bm}

\def\eqref#1{equation~\ref{#1}}

\def\1{\bm{1}}

\DeclareMathAlphabet{\mathsfit}{\encodingdefault}{\sfdefault}{m}{sl}
\SetMathAlphabet{\mathsfit}{bold}{\encodingdefault}{\sfdefault}{bx}{n}

\usepackage{hyperref}
\usepackage{url}
\usepackage{mdframed}
\usepackage{amsmath}
\usepackage{extpfeil}
\usepackage{wrapfig}
\usepackage{enumitem}
\usepackage{url}

\title{Multi-Modal Forecaster: Jointly Predicting Time Series and Textual Data}

\author{
\textbf{Kai Kim}$^{1}$, \textbf{Howard Tsai}$^{1}$, \textbf{Rajat Sen}$^{2}$, \textbf{Abhimanyu Das}$^{2}$, \\
\textbf{Zihao Zhou}$^{1}$, \textbf{Abhishek Tanpure}$^{1}$, \textbf{Mathew Luo}$^{1}$, \textbf{Rose Yu}$^{1}$ \\
$^{1}$Department of Computer Science, University of California, San Diego \\
\texttt{\{mik009, cht028, ziz244, atanpure, mal084, roseyu\}@ucsd.edu} \\
$^{2}$Google Research \\
\texttt{\{senrajat, abhidas\}@google.com}
}

\begin{document}

\maketitle

\begin{abstract}
Current forecasting approaches are largely unimodal and ignore the rich textual data that often accompany the time series due to lack of well-curated multimodal benchmark dataset.
In this work, we develop TimeText Corpus (TTC), a carefully curated, time-aligned  text and time dataset for multimodal forecasting.
Our dataset is composed of sequences of numbers and text aligned to timestamps, and includes data from two different domains: climate science and healthcare.
Our data is a significant contribution to the rare selection of
available multimodal datasets. We also propose the Hybrid Multi-Modal Forecaster (Hybrid-MMF), a
multimodal LLM  that jointly forecasts both text and time series data using
shared embeddings.
However, contrary to our expectations, our Hybrid-MMF model does not outperform existing baselines in our experiments. This negative result highlights the challenges inherent in multimodal forecasting. Our code and data are available at {\url{https://github.com/Rose-STL-Lab/Multimodal_Forecasting}}.

\end{abstract}

\section{Introduction}\label{sec:intro}
Deep learning has become the predominant method in
forecasting large-scale time series \cite{zhou2022fedformer,wang2022koopman, woo2023learning}, but most existing methods consider time series as a
single data modality. In practice, time series data do not exist in isolation and there are rich text meta-data available. Large Language Models (LLM) such as GPT \cite{brown2020language} and LLaMA \cite{touvron2023llama} have demonstrated marvelous success in processing and understanding text data. It is intriguing to exploit text as an new  modality to improve forecasting. On one hand, text can provide
context to the dynamics governing the associated time series. On the other hand, generating text alongside numerical forecasts also offer valuable interpretations for the predictions. 

Effectively combining and forecasting these two data modalities remains a complex challenge \cite{zhang2024llmstime}. The differences in structure and content between numerical time series and text data pose challenges to their simultaneous integration in forecasting models. Traditionally, the community has studied time series and natural language processing separately. To the best of our knowledge, there is no large-scale, well-curated, paired time series and text dataset for forecasting. The only exception is Time-MMD \cite{liu2024timemmd}, a curated dataset containing valuable information for numerical forecasting across multiple domains. Notably, Time-MMD's text data is processed from real-world sources using models such as Llama 70B to ensure meaningful and contamination-free content. While this dataset offers useful insights and has shown to be useful for accurate time-series forecasting \cite{jin2024timellm, cao2024tempo}, our preliminary fine-tuning of LLMs on Time-MMD for text prediction highlighted the need for further refinement and expanded data collection to improve performance on text prediction.

Several early attempts have been made at multimodal forecasting. \cite{kumar2022multimodal, obst2019textual} propose to augment time series forecasting with textual meta-data, but only predict time series. Other works proposed to treat time series forecasting as a ``modified language task''. For example, LLMTime \cite{gruver2023large} proposes a zero-shot approach by representing the time series as a string of numerical digits inputs to LLMs.  Time-LLM \cite{jin2023time} transforms the time series input into text prototype representation and projects the output back to numerical forecasts.  Others use a pre-trained language model with fine-tuning or prompt engineering for forecasting \cite{zhou2023one, cao2023tempo}. While these approaches show promise in leveraging LLMs for time series forecasting, they primarily focus on fitting numerical data into the language model paradigm rather than addressing the broader challenge of multimodal forecasting.

In this work, we focus on the challenge of joint multimodal forecasting—forecasting both time series and textual event data simultaneously. To address this, we introduce the TimeText Corpus (TTC), a carefully curated, time-aligned text and time dataset for multimodal forecasting. Our dataset is composed of sequences of numbers and text aligned to timestamps, and includes data from two different domains: climate science and healthcare. Additionally, we propose a hybrid approach to model this data by integrating both modalities, aiming to exploit the complementary nature of numerical and textual data for better forecasts.

Our contributions include: \begin{itemize} \item \textbf{Simultaneous Encoding of Multimodal Data:} We propose and experiment with techniques for jointly encoding numerical and textual data into shared embeddings for simultaneous multimodal forecasting. \item \textbf{Multimodal Dataset Contributions:} We curate two datasets: one from Mimic-III (medical), and the other from the National Weather Service (climate), which can serve as benchmarks for future research. \end{itemize}

In the following sections, we detail our TimeText Corpus (TTC) and propose the Hybrid Multi-Modal Forecaster (Hybrid-MMF), a model that jointly forecasts both text and time series data. Our experiments show that while our model demonstrates competitive performance, marginal improvements were observed over baselines in several cases.

\section{Related Work}\label{sec:related}

\paragraph{Traditional Time Series Forecasting.} 
Classical time series models such as AR, ARIMA, the VAR models~\citep{zivot2006vector} and exponential smoothing models~\citep{mckenzie1984general} rely on over-simplified modeling assumptions and often cannot cope with the complexity of real-world data. The rise of deep learning has shown great success in forecasting time series at scale. For example,  DeepAR ~\citep{salinas2020deepar}, Koopman Neural Forecaster ~\citep{wang2022koopman} and PatchTST \citep{nie2022time} are all recent models that meet
or exceed traditional ARIMA and ETS forecasting methods on many data domains.

\paragraph{Deep Learning and Transformers.}
MLPs have been used for time-series forecasting in the popular N-BEATS model~\citep{oreshkinn}. TiDE \citep{das2023longterm} uses a Multi-layer Perceptron (MLP)-based encoder-decoder framework. Gated recurrent units like GRU and LSTM
were introduced to handle the autoregressive nature of time-series data,
though these methods are typically data hungry and have complex parameterizations \citep{parmezan2023timesota}.

Transformers have recently seen a surge in popularity for time series forecasting. LongTrans~\citep{li2019enhancing} uses an attention layer with LogSparse design, reaching near-linear space and time complexity. Informer~\citep{zhou2021informer} uses the ProbSparse self-attention mechanism to achieve sub-quadratic dependency on the length of the context. Autoformer~\citep{wu2021autoformer} employs trend and seasonal decomposition with a sub-quadratic self-attention mechanism. FEDFormer~\citep{zhou2022fedformer} incorporates a frequency-enhanced structure, while Pyraformer~\citep{liu2021pyraformer} introduces pyramidal self-attention that achieves linear complexity and can attend to different granularities.

Transformer architectures must be parameterized well in order to 
be effective~\citep{nie2022time}. \citet{zeng2022transformers} recently observed that a simple 
linear projection model can outperform transformer based architectures on 
long horizon time series forecasting tasks.
DLinear~\citep{zeng2022transformers} learns a linear mapping from context to the horizon, pointing to deficiencies in sub-quadratic approximations to the self-attention mechanism. PatchTST~\citep{nie2022time} resolved this by showing 
that feeding contiguous patches of time-series as tokens to the vanilla self-attention mechanism can beat the performance of DLinear in long-term forecasting benchmarks.

All of the works mentioned above only use a single time series modality and benchmark on time series datasets that are mostly stationary. Our work builds
on the success of deep learning methods by building a multi-modal model that forecasts
both text and time. 

\paragraph{LLM and Time Series.} 
The use of Large Language Models (LLMs) as natural time-series predictors have
been explored in varying levels of sophistication, from zero-shot forecasting 
to incorporating time series embeddings in various levels within the language model~\citet{zhang2024llmstime}. 

\emph{Prompt-based methods.} LLMTime \citep{gruver2023large} uses LLM for time series forecasting in a zero-shot manner. They represent the time series as a string
of numerical digits and view time series forecasting as next-token prediction in text.  Time-LLM ~\citep{jin2023time} proposes to reprogram pre-existing LLMs by translating time series data into a text-like format. ~\citet{zhou2023one} use pre-trained Large Models from vision and language to fine-tune on all major types of tasks
involving time series. \citet{chang2024eventforecasting} forecast textual events,
rather than time series data, using a RAG-based method. 

\emph{Quantization and Alignment.} Quantization based methods convert numerical data into 
discrete representations for input to LLMs, whereas alignment-based methods train a 
separate encoder for time series data and align the encoded time series to the 
semantic space of language models ~\citep{zhou2023one}.
Many different models across different domains adopt one or both of these techniques.
Auto-TTE \citep{chung2023textecg} quantizes ECG graphs into
discrete formats and forecasts ECG signals conditioned on text reports. 
~\citet{cao2023tempo} propose TEMPO, a novel
integration of seasonal and trend decomposition into the pre-trained transformers. 

Overall, these techniques focus on using or tuning pre-trained LLMs for time-series
forecasting. We build on these methods by predicting text as an additional modality,
and use techniques from papers like LLMTime \citep{gruver2023large} to benchmark
our results. 

\paragraph{Multimodal Forecasting.}
This work is the first investigation of a multimodal forecasting model that incorporates textual and numerical time-series data. Previous works have mainly focused on study in specific domains and single modality. 
\citet{kumar2022multimodal} explored a multi-modal sales forecasting network that combines data from news articles with numerical data such as historical sales and holiday information. 
\citet{obst2019textual} leveraged daily weather reports to predict time series of national electricity consumption, average temperature, and wind-speed.  

Recently, \citet{liu2024timemmd} jointly proposed Time-MMD, a new multimodal dataset,
and MM-TSFlib, a new multimodal forecasting model.
MM-TSFlib, forecasts time series by first modeling the time and textual series
independently, then combining the outputs using a linear weighting mechanism. 
techniques differ from ours in two important ways. First, their time series
pipeline relies on a frozen LLM to generate the textual embeddings~\citep{liu2024timemmd}, 
whereas our hybrid model can learn the textual embeddings.
Second, Time-MMD does not study
how multi-modal inputs can be used to create textual predictions, which we show
beats naive text-only prediction models and can serve as a powerful tool for event
forecasting.
\section{Dataset}\label{sec:dataset}

\paragraph{Weather.}
National weather service dataset \footnote{\url{https://www.wpc.ncep.noaa.gov/archives/web_pages/medr/get_medr_products_bck.php}} consists of 2 separate sources of data: a series of text discussions about weather forecasting from the National Weather Service and a series of numerical weather recordings from Visual Crossing. Using the National Weather Service archives, we obtain daily forecast discussions about the different regions across the United States. In conjunction with the text data, we collect daily weather recordings in Washington D.C. by Visual Crossing Weather API. The finalized dataset, which includes weather recordings and forecast discussions, ranges from Jan 1st, 2014 to December 1st, 2023, for a total of 3621 days. The time series categories include daily maximum temperature, minimum temperature, average temperature, dew point, humidity, precipitation, and average windspeed. Different time series on the same day are paired with the same text data. We use these daily weather recordings to forecast the future values in Washington D.C., with forecasting horizons of 1 days through 7 days.

\paragraph{Medical.}
MIMIC-III\footnote{\url{https://physionet.org/content/mimiciii/1.4/}} comprises of hourly data from over 58,000 hospital admissions of more than 38,000 patients. It spans a period of 12 years, from 2001 to 2012, enabling researchers to study healthcare trends and seasonalities over time. This dataset includes a mix of structured data such as lab results, vital signs, and medications, along with unstructured data like clinical notes. We selected patients and items based on the pairs with the highest number of associated measurements. Some of the sequences contain missing data. To reduce the impact of these gaps, we filtered out data with more than a 5\% rate of missing days. This resulted in a dataset of 50 time series of different patients and vital signs, with each time series spanning 3 to 5 months and a total of 7,589 daily timesteps. Each of the time series corresponds to one patient and one of the following vital signs: Heart Rate, Respiratory Rate, SaO2, and FiO2. The nurse's notes from the MIMIC dataset provide detailed information on the patient's vital signs and important findings from the tests performed. Because the notes are already recorded daily and do not exceed the input limit, we do not summarize the text.

\begin{wraptable}{r}{0.4\textwidth}
\caption{Comparison of dataset sizes in terms of time steps.}
\begin{tabular}{lcc}
\toprule
Dataset & Time steps & Frequency\\
\midrule
Weather & 3,621 & 1 day\\
Medical & 7,589 & 1 day\\
\bottomrule
\end{tabular}

\label{tab:dataset-comparison}
\end{wraptable}

\paragraph{Pre-Processing}

We use GPT4o-mini to parse and extract relevant information from the raw unstructured dataset into categories related to the time series variables. For the weather forecast dataset, we extract text relevant to temperature, humidity, precipitation, and wind speed. For the medical dataset, we extract text relevant to heart rate, respiratory rate, SaO2, and FiO2. We extract relevant information by first chunking each text into 1000 letters. Then, we refine the summarization by providing the original raw data to fill in incomplete or inaccurate information. Finally, the chunk summaries are combined with our specific prompts (appendix). The final dataset contains 10 years of daily weather forecast and average 105 daily for each of the 74 patients for the medical dataset. Table~\ref{tab:dataset-comparison} shows the number of time steps in each dataset.

\section{Method}\label{sec:method}
We present Hybrid Multi-Modal Forecaster (Hybrid-MMF), a novel framework to enhance both text and time series forecasting accuracy by encoding both historial time series and correlated text and projecting them into different output heads.

\subsection{Variable Definitions}
These variables are used in the figure presentation below.

\begin{itemize}
    \item \( I \): Input window size or number of time steps.
    \item \( C \): Number of input time-series variables.
    \item \( N \): Length of tokenized text per time step.
    \item \( E \): Embedding dimension for the text data.
    \item \( H \): Hidden dimension after projection.
    \item \( O \): Number of output time-series variables or text tokens.
    \item \( V \): Vocabulary size for LLM tokenization.
\end{itemize}

\subsection{Problem Definition}

Given the time series dataset of length $T$, denoted as $X = [X_1, \cdots, X_T]$ with each $X_t | _{t \in [1, T]} \in \mathbb{R}^d$ being a state vector at time $t$, as well as the text data $Y= [Y_1,\cdots, Y_T]$ with $Y_t$ representing the text corpus collected at time $t$, our goal is to learn a functional mapping $f$ such that 
\begin{equation}
[X_{t-L}, \cdots, X_{t-1}, Y_{t-L}, \cdots, Y_{t-1}] \xlongrightarrow[]{f}[X_{t-L}, \cdots, X_{t-1}, Y_{t-L}, \cdots, Y_{t-1}] \nonumber
    \label{eqn:def}
\end{equation}
where $L$ is the input sequence's time lag and $H$ is the forecasting horizon. $L$ is chosen to ensure the temporal dependencies and trends in the input sequences are considered. $H$ is task-dependent.

\subsection{Hybrid Model}

Our hybrid model leverages Llama 3.1 8B, a state-of-the-art pre-trained language model, to embed text inputs. Llama 3.1 8B was chosen for its robust capability in handling complex natural language processing tasks, especially in generating high-quality embeddings that capture deep semantic and contextual relationships. This allows our model to understand and predict textual events with high precision.

For numerical inputs, we project the data into the textual embedding space using either the Llama 3.1 8B model or the BAAI/bge-small-en-v1.5 model. The motivation behind this choice is that both models are highly effective in generating embeddings that can represent multiple modalities. By chosing these embeddings, the model can effectively handle the multimodal nature of the forecasting task, predicting both time-series data and textual events seamlessly. The final combined embedding forms a comprehensive representation, enhancing the model's ability to capture dependencies across different modalities. The architecture of the model is shown in full below in Figure \ref{fig:hybrid_model}.

\begin{figure}[htbp]
    \centering
    \includegraphics[width=1\textwidth]{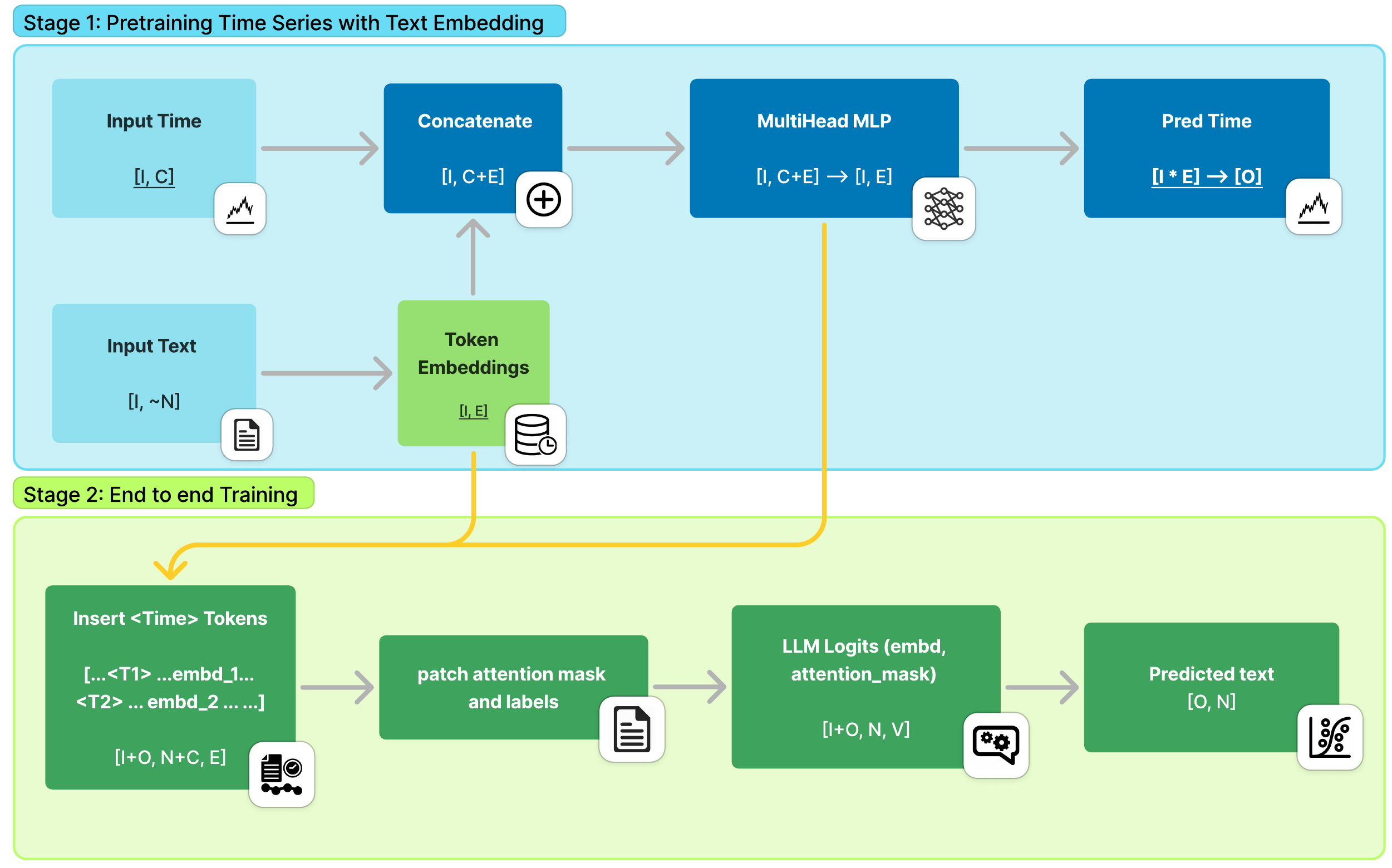}
\caption{The diagram shows the architecture of the Hybrid Multi-Modal Forecaster (Hybrid-MMF) for integrating time series and text data in two stages. In \textbf{Stage 1 (Pretraining)}, the model processes (1) time series data $[I, C]$ and text data $[I, \sim N]$ by embedding and concatenating them [I, E]. A multi-head MLP $[I, C+E] \rightarrow [I, E]$ creates a hidden state that is used to predict time series values $[I \cdot E] \rightarrow [O]$. In \textbf{Stage 2 (End-to-End Training)}, token embeddings $[I, N, E]$ are combined with stage 1's hidden state at the beginning of each timestep, and passed through a large language model (LLM). The model predicts both text and time series outputs, with logits $[I + O, N, V]$ producing the final text sequence $[O, N]$.}
    \label{fig:hybrid_model}
    \vspace{-3mm}
\end{figure}

\subsection{Stage 1: Pretraining Time-Series with Text Embeddings}
In the first stage, the model processes two types of inputs:

\textbf{Input Time Series:} \( \mathbf{X}_{time} \in \mathbb{R}^{I \times C} \), where \( I \) is the number of time steps, and \( C \) represents the number of variables or channels in the time-series data.

\textbf{Tokenized Input Text:} \( \mathbf{X}_{text} \in \mathbb{R}^{I \times N} \), where \( N \) is the tokenized text length per time step.

The text data is converted into embeddings \( \mathbf{E}_{text} \in \mathbb{R}^{I \times E} \) using either the LLaMA tokenizer or the \textit{BAAI/bge-small-en} model from Sentence Transformers. The text embeddings are projected to match the hidden dimension \( H \), which allows for concatenation with the time-series data, resulting in a combined representation \( \mathbf{X}_{concat} \in \mathbb{R}^{I \times (C+H)} \).

This concatenated data is then fed through a multihead MLP module, which transforms it into hidden states that encode the information from both time-series and text inputs. The model outputs predictions \( \mathbf{\hat{Y}}_{time} \in \mathbb{R}^{I \times O} \), where \( O \) is the number of output time-series variables.

\subsection{Stage 2: Full End-to-End Training}

In this stage, the model is jointly trained by integrating the hidden states from the multihead MLP with an LLM module.

\textbf{Tokenized Text:} The input text is tokenized into \( \mathbf{T}_{emb} \in \mathbb{R}^{I \times N \times V} \), where \( N \) is the token length, and \( V \) is the vocabulary size.

\textbf{Hidden State Insertion:} The hidden states from the multihead MLP (\( \mathbf{H} \in \mathbb{R}^{I \times H} \)) are concatenated with the tokenized text for each input window. This step allows the LLM to process both time-series and text data simultaneously.

The temporal dependencies are modeled by feeding attention masks and labels into the LLM at each time step. The LLM then applies attention mechanisms across both modalities to generate final predictions \( \mathbf{\hat{Y}}_{final} \in \mathbb{R}^{O \times N} \), which can be used for either time-series forecasting or text generation.


The entire model is optimized using backpropagation, employing a custom loss function. For time-series forecasting, we apply the Mean Squared Error (MSE) loss, while Cross-Entropy Loss is used for text prediction.

\section{Experiments}\label{sec:experiments}

We conducted a series of experiments to evaluate our proposed 
hybrid model against many text-only and numerical-only
forecasting models, which served as our baseline.  A full explanation of the evaluated
models can be found in \ref{app:models_metrics}.  The textual
metrics and numerical metrics were evaluated independently. 
A complete collection of the results can be found in 
Appendix \ref{app:quant_results}.

\subsection{Baseline Models}
We implement the following baseline models for comparison. For language model-based approaches (Text2Text, TextTime2Text, TextTime2Time, and TextTime2TextTime), we use the unsloth/Meta-Llama-3.1-8B-Instruc model under zero-shot, in-context learning (with one example), and fine-tuned settings. The context length varies from 900 to 5500 tokens depending on the input-output prediction days and specific dataset characteristics. All other baselines are evaluated in zero-shot settings only.
\begin{itemize}[leftmargin=*]
\item \textbf{Input\_Copy}: A simple baseline that copies the input as the output, serving as a performance floor for both text and time-series predictions.
\item \textbf{NLinear}: A linear projection time series forecasting model proposed by \citep{zeng2022transformers}, focused solely on numerical predictions.

\item \textbf{Linear\_Text\_Embedding}: An extension of the NLinear model that incorporates text embeddings as covariates, using the \textit{bge-small-en-v1.5} embedding model.

\item \textbf{Text2Text}: Language models that process and forecast textual events only using unsloth/Meta-Llama-3.1-8B-Instruc.

\item \textbf{TextTime2Text}: Language models that take both text and time-series inputs (with time-series data converted to text) but forecast only textual events.

\item \textbf{TextTime2Time}: Language models that process both text and time-series inputs but predict only future time-series values. The textual input provides context for numerical forecasting.

\item \textbf{TextTime2TextTime}: Language models that handle both modalities for input and output, where numerical data is stringified and concatenated with text. Capable of predicting both text and time-series data.

\item \textbf{PatchTST}: A Patch-based Time Series Transformer model for forecasting, as proposed by \citep{nie2022time}.
\end{itemize}

\subsection{Experiment Setup}\label{subsec:metrics}

We employed different evaluation metrics for text and time-series predictions. For time-series predictions, we used Root Mean Squared Error (RMSE).
To evaluate the forecast text, we use the following metrics.
\begin{itemize}[leftmargin=*]
    \item \textbf{Cosine Similarity:} We calculated cosine similarity as the dot product between the sentence embeddings of the generated and reference texts.
    \item \textbf{METEOR:} METEOR \citep{banerjee-lavie-2005-meteor} is a metric  that considers synonyms, stemming, and paraphrases when comparing
    generated and reference texts.
    \item \textbf{ROUGE:} ROUGE \citep{lin-2004-rouge} is a family of 
    metrics which assess similarity by comparing the number of overlapping
    n-grams.
    \item \textbf{GPT Semantic Score:} We ask ChatGPT-4o to rank the similarity between the 
    the reference text and the predicted text using a scale from
    1-10.
    \item \textbf{GPT F1 Score:} As a last textual evaluation metric, we prompt ChatGPT-4o to first compare line by line the ground truth and prediction text. Then, we count up the number of true positive, false positive, and false negative facts to compute the Precision, Recall, and F1 scores. Full details can be found in Appendix \ref{app:experimental_details}.
\end{itemize}

In our forecasting setting, the time lag is the same as forecasting horizon. For Climate, we forecast up to 7 days and (1-1) means 1 day ahead forecasting with lag 1. For Medical, we forecast 6 time steps ahead.  We tested the model's performance across increasing  horizons for all tasks.
Our experiments are based on \textit{univariate} forecasting. We chose patient heartrate 
and temperature in Washington D.C. as our forecasting variables for the 
climate and medical dataset, respectively.

\subsection{Quantitative Results}
Table~\ref{tab:climate-rmse} and Table~\ref{tab:medical-rmse} shows the RMSE error of number prediction.
Table~\ref{tab:weather-forecast-comparison} visualize the text prediction for climate dataset 1-1 case from our fine-tuned model. The best values in each column are presented in bold,
and the second best values are underlined. The textual
forecasting results can be found in Appendix 
\ref{app:quant_results}.

\begin{table}[htbp!]
  \caption{RMSE for various models on \textbf{Weather} dataset with different window sizes. We include an extra
  7-7 forecast to record all of our experimental results for completeness. }
  \label{tab:climate-rmse}
  \centering
  \small
  \vspace{0.5cm}
  \begin{tabular}{l|ccccccc}
    \toprule
    \textbf{Model} & \textbf{1-1} & \textbf{2-2} & \textbf{3-3} & \textbf{4-4} & \textbf{5-5} & \textbf{6-6} & \textbf{7-7} \\
    \midrule
    Input\_Copy & 5.027 & 7.222 & 8.133 & 8.504 & 8.692 & 8.839 & 8.834 \\
    NLinear & 4.981 & 6.129 & 6.501 & 6.710 & 6.834 & 6.916 & 6.962 \\
    NLinear\_Text\_Embedding & \underline{4.835} & 5.800 & \underline{5.951} & \textbf{5.934} & \textbf{6.022} & \textbf{6.024} & \textbf{6.106} \\
    TextTime2TextTime (zero-shot) & 7.150 & 7.818 & 9.842 & 9.782 & 9.693 & 9.679 & 9.804 \\
    TextTime2Time (zero-shot) & 6.400 & 7.936 & 9.253 & 9.984 & 10.179 & 9.880 & 10.100 \\
    TextTime2TextTime (in-context) & 7.092 & 8.141 & 8.437 & 8.743 & 8.574 & 9.156 & 9.241 \\
    TextTime2Time (in-context) & 7.215 & 10.550 & 10.509 & 9.548 & 9.436 & 10.094 & 9.161 \\
    TextTime2TextTime (fine-tuned) & 5.243 & 5.955 & 6.724 & 7.253 & 7.678 & 8.034 & 7.666 \\
    TextTime2Time (fine-tuned) & 5.268 & 6.323 & 7.139 & 7.736 & 8.131 & 7.908 & 7.849 \\
    PatchTST & 4.912 & \textbf{5.305} & 6.021 & 6.576 & 6.980 & 7.170 & 7.360 \\
    Hybrid (pretrain MLP + Full) & \textbf{4.759} & \underline{5.597} & \textbf{5.906} & \underline{6.019} & \underline{6.133} & \underline{6.027} & \underline{6.143} \\
    Hybrid (pretrain mlp\&L1M + full) & 4.888 & 5.689 & 5.960 & 6.168 & 6.461 & 6.543 & 6.647 \\
    \bottomrule
  \end{tabular}
\end{table}
\vspace{-6mm}

\begin{table}[H]
  \caption{RMSE for various models on \textbf{Medical} dataset with different window sizes}
  \label{tab:medical-rmse}
  \centering
  \small
  \vspace{0.5cm}
  \begin{tabular}{l|cccccc}
    \toprule
    \textbf{Model} & \textbf{1-1} & \textbf{2-2} & \textbf{3-3} & \textbf{4-4} & \textbf{5-5} & \textbf{6-6}\\
    \midrule
    Input\_Copy & 5.235 & 5.909 & 6.084 & 6.544 & 7.094 & 7.479 \\
    NLinear & \underline{5.195} & \underline{5.279} & \underline{5.275} & \underline{5.406} & \underline{5.562} & \underline{5.875} \\
    NLinear\_Text\_Embedding & \textbf{5.117} & \textbf{5.143} & \textbf{5.106} & \textbf{5.300} & \textbf{5.492} & \textbf{5.759} \\
    TextTime2TextTime (zero-shot) & 11.549 & 8.664 & 8.388 & 9.057 & 8.976 & 8.963 \\
    TextTime2Time (zero-shot) & 6.706 & 8.000 & 7.529 & 7.179 & 7.935 & 8.151 \\
    TextTime2TextTime (in-context) & 6.600 & 6.620 & 6.521 & 7.291 & 7.967 & 8.165 \\
    TextTime2Time (in-context) & 7.182 & 6.414 & 6.479 & 7.226 & 7.527 & 7.868 \\
    TextTime2TextTime (fine-tuned) & 6.689 & 6.432 & 6.022 & 6.483 & 6.747 & 6.731 \\
    TextTime2Time (fine-tuned) & 6.084 & 6.246 & 6.202 & 6.709 & 7.298 & 7.157 \\
    PatchTST & 5.735 & 6.757 & 7.350 & 7.687 & 7.996 & 8.470 \\
    Hybrid w/ llama tokens & 5.202 & 5.472 & 6.620 & 6.269 & 8.673 & 8.454 \\
    Hybrid w/ bge embeddings & 11.657 & 9.054 & 5.776 & 8.633 & 8.625 & 8.694 \\
    \bottomrule
  \end{tabular}
  \vspace{-3mm}
\end{table}

\begin{table}[ht!]
\centering
\small
\begin{tabular}{p{0.15\textwidth}p{0.25\textwidth}p{0.25\textwidth}p{0.25\textwidth}}
\toprule
\textbf{Feature} & \textbf{Day 1 (Input)} & \textbf{Day 2 (Prediction)} & \textbf{Day 2 (Ground truth)}\\
\midrule
Date & 2022-12-31 & 2023-01-01 & 2023-01-01 \\
\midrule
Temperature & 51.2°F & 44.9°F & 51.8°F\\
\midrule
Weather Forecast & Winter storm threat across the upper Midwest on Tuesday with heavy snow and increasingly windy conditions expected. Milder than average temperatures, with lows 15-30°F above normal and highs 10-20°F above normal in the eastern U.S., reaching 20-30°F anomalies in the Ohio Valley and Northeast. Significant heavy rain and severe weather threats across the Southeast Tuesday into Wednesday, particularly in the Florida Panhandle. Additional rounds of moderate to heavy precipitation forecast for the West, primarily in California, with potential Atmospheric River impacts and multi-day heavy snow threats in the Sierra. There is uncertainty about the deep cyclone's formation and track off the Pacific Northwest coast, with trends toward increased precipitation amounts as the week progresses. & Heavy rain and severe weather expected across the Southeast and Mid-Atlantic on Tuesday and Wednesday, with a winter storm threat in the upper Midwest and Great Lakes. A deep cyclone will develop near the Pacific Northwest coast, leading to increased precipitation and strong winds in the Northwest and northern Rockies. Milder than average temperatures will persist in the East, with highs 10-20°F above normal, while the West will experience a cooling trend with highs 10-20°F below normal. The Southeast will see a significant rainfall threat, with potential for severe weather, particularly in Florida. Heavy rain is also expected in California, with potential Atmospheric River (AR) impacts, while the Sierra Nevada will face multi-day heavy snow threats. A deep cyclone may develop near the Pacific Northwest coast, with uncertainty regarding its track and strength. & Heavy precipitation is expected in California, particularly peaking Wednesday-Thursday, associated with a deep cyclone bringing moisture. A faster-moving cold front is forecast for the eastern U.S., resulting in a decreased threat of heavy rain in the Southeast and colder air from eastern Canada impacting temperatures Thursday and Friday. Snow or mixed precipitation may occur in the interior Northeast due to a sheared low-pressure wave and lingering upper wave interaction. Moderate rain is expected along the Eastern Seaboard midweek from a wavy frontal system. Much milder than average temperatures, 15-30\u00b0F above normal, will persist in the eastern U.S., with record warm morning lows on Wednesday. Additional precipitation rounds will reach the West Coast, with a potential Atmospheric River event and a slight risk of excessive rainfall. Heavy snow is anticipated in the Sierra, Cascades, and into the Intermountain West. Next weekend may see limited QPF potential as systems translate across the central to east-central U.S.\\
\bottomrule
\end{tabular}
\caption{Example of weather forecast input data (Day1), subsequent prediction (Day2) and ground-truth(Day2).}
\label{tab:weather-forecast-comparison}
\end{table}

\begin{figure}[ht!] 
\centering 
\includegraphics[width=\textwidth]{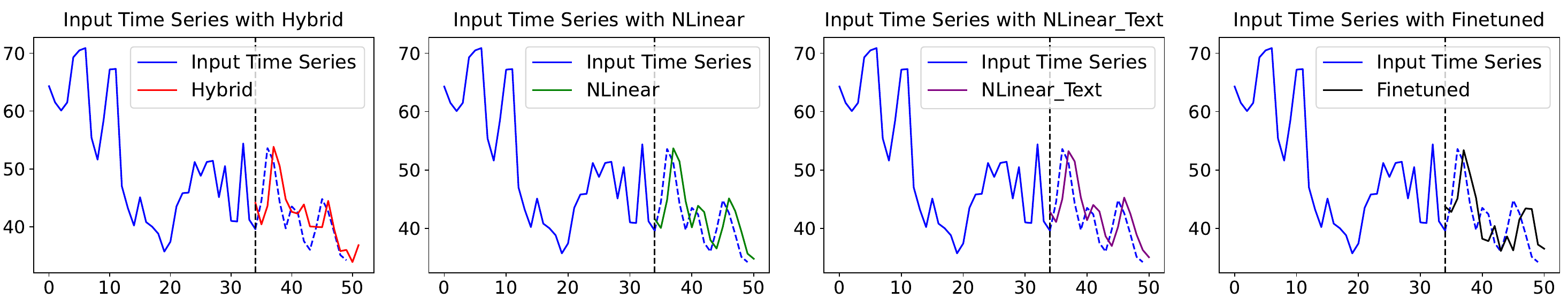} 
\caption{Comparison of 1-1 predictions across models on the weather dataset. The blue line denotes the input time series, and the colored lines correspond to the predictions from each model.} 
\label{fig:climate1-1} 
\vspace{-3mm}
\end{figure}

\subsection{Weather Dataset Prediction Visualization} We present a visualization comparing the predictions of four models—Hybrid, NLinear, NLinear with text embeddings, and Finetuned—using the 1-day input to 1-day output (1-1) setup on the climate dataset. As shown in Figure \ref{fig:climate1-1}, the blue line represents the input time series, while each model's predictions are depicted in different colors. The purely textual method of using a finetuned model for prediction is qualitatively more inaccurate than pure time models (NLinear) and the hybrid text-time models (Hybrid and NLinear-Text).

\section{Discussion}\label{sec:discussion}
In this study, we explored different strategies for multimodal forecasting by combining both text and numerical data to improve prediction accuracy. Our results showed that the fine-tuned model consistently outperformed both the few-shot and zero-shot settings. Our hybrid model matched but did not
exceed these baselines. We conjecture
that this is due to the inherent difficulty in improving on
textual forecasting and the inability for models to adapt
well to the time-series token in the model's embedding space.

We also found that incorporating text embeddings improved the numerical predictions in both the NLinear and fine-tuned models. This suggests that textual context can help enhance the model’s ability to recognize patterns in numerical data. However, despite these improvements, our hybrid model—which was designed to forecast both text and numbers—did not meet expectations and failed to surpass the NLinear baseline.

A significant challenge we encountered is the inherent difficulty of textual event forecasting. Taking example
from our data domains, 
tomorrow’s weather is a highly dynamic, unpredictable 
system. The data we used may not have been large or varied enough to enable the hybrid model to learn effectively.

In conclusion, while the fine-tuned model and the addition of text embeddings led to clear improvements, the hybrid model didn’t achieve the breakthrough we were aiming for. This highlights the complexity of predicting future outcomes and underscores the need for larger datasets and further research to fully tap into the potential of multimodal forecasting.

\section{Acknowledgment}
This work was supported in part by the U.S. Army Research Office
under Army-ECASE award W911NF-07-R-0003-03, the U.S. Department Of Energy, Office of Science, IARPA HAYSTAC Program, and NSF Grants \#2205093, \#2146343, \#2134274, CDC-RFA-FT-23-0069,  DARPA AIE FoundSci and DARPA YFA.

\newpage
\bibliography{iclr2025_conference}
\bibliographystyle{iclr2025_conference}

\appendix
\newpage
\section{Additional Experimental Details}\label{app:experimental_details}
\subsection{LLM Finetunes}
We fine-tune the unsloth/Meta-Llama-3.1-8B-Instruct model using Low-Rank Adaptation (LoRA) \citep{hu2021lora} to improve computational efficiency. LoRA allows us to update only specific layers of the model, such as the attention and feed-forward layers, while freezing the remaining parameters. This approach significantly reduces the memory and computational cost associated with traditional fine-tuning.

The text data is tokenized using the LLaMA tokenizer \citep{touvron2023llama}, while the numerical data is fed directly into the model without normalization. During fine-tuning, the model learns to predict both text and numerical outputs, effectively handling multimodal input. This approach allows the model to generate coherent text and accurate numerical predictions, which are subsequently compared to baseline models.

\subsection{Few-shot and Zero-shot Evaluation}

In our few-shot learning setup, the model is given only one random input-output example from training dataset. This approach allows us to evaluate the model's ability to generalize with minimal data, which is crucial in real-world scenarios where labeled data may be scarce. By measuring the performance on such limited examples, we can assess the model's adaptability and effectiveness in low-data environments.

For zero-shot evaluation, the model is tested without any task-specific fine-tuning or exposure to the target data. This setting evaluates the pre-trained model's ability to generalize to entirely new tasks based solely on its prior knowledge. Zero-shot results provide insight into the model’s flexibility and robustness, which is especially valuable when no task-specific data is available for model training or fine-tuning.
\newpage
\section{Prompt}\label{app:prompt}
\subsection{Prompt for weather dataset}
\begin{mdframed}[linewidth=1pt, roundcorner=5pt]
\noindent
\textbf{\large{Prompt:}} \\
Given the weather information of the first 1 day, predict the weather information of the next 1 day. Output the result strictly in the following JSON format and no additional text:

\begin{verbatim}
{
    "day_2_date": "YYYY-MM-DD",
    "day_2_weather_forecast": "Weather description",
    "day_2_temp": "A Float Number"
}
\end{verbatim}

Input
\begin{verbatim}
{
    "day_1_date": "2014-01-05",
    "day_1_weather_forecast": "{weather information}",
    "day_1_temp": 37.1
}
\end{verbatim}

\vspace{0.5cm}

\textbf{\large{Model Response:}}
\begin{verbatim}
{
    "day_2_date": "2014-01-06",
    "day_2_weather_forecast": "weather information",
    "day_2_temp": 36.6
}
\end{verbatim}
\end{mdframed}

\subsection{Prompt for medical dataset}

\begin{mdframed}[linewidth=1pt, roundcorner=5pt]
\noindent
\textbf{\large{Prompt:}} \\
Given the medical information of the first 1 day, predict the medical information of the next 1 day. Output the result strictly in the following JSON format and no additional text:
\begin{verbatim}
{
    "day_2_date": "YYYY-MM-DD",
    "day_2_medical_notes": "Medical description",
    "day_2_Heart_Rate": "A Float Number"
}
\end{verbatim}

Input:
\begin{verbatim}
{
    "day_1_date": "2131-05-06",
    "day_1_medical_notes": "{medical information}",
    "day_1_Heart_Rate": 170.5
}
\end{verbatim}

\textbf{\large{Model Response:}}
\begin{verbatim}
{
    "day_2_date": "2131-05-07",
    "day_2_medical_notes": "medical information",
    "day_2_Heart_Rate": 169.815
}
\end{verbatim}
\end{mdframed}

\subsection{Prompt for GPT Score}
\begin{mdframed}[linewidth=1pt, roundcorner=5pt]
\noindent
\textbf{\large{Prompt:}} \\
You are a helpful assistant capable of evaluating the semantic similarity between two summaries. 
The semantic score you provide should be a number between 1 and 10, where 10 represents the highest level of semantic similarity 
(meaning the summaries convey almost identical information), and 1 represents the lowest level of semantic similarity (meaning the summaries convey entirely different or unrelated information). 
The score should reflect how closely the meanings and key details of the two summaries align. You should only give me the number, nothing else. \\
Ground truth: \{ground\_truth\} \\
Output: \{output\} \\

\textbf{\large{Model Response:}}
\begin{verbatim}
{
    "score": 5
}
\end{verbatim}
\end{mdframed}

\subsection{Prompt for GPT4 text evaluator}
\begin{mdframed}[linewidth=1pt, roundcorner=5pt]
\noindent
\textbf{\large{Prompt:}} \\
You are a helpful assistant capable of evaluating the similarity, accuracy, and consistency of a ground truth weather forecast, and predicted weather forecast for the same day.
By comparing the semantic meaning of each sentences line by line, count the number of \\
1) (True Positive): Correct information presented in the prediction, \\
2) (False Postive): Incorrect information presented in the prediction, \\
3) (False Negative): Information not presented in the prediction. \\
Make sure to include explanation for each step.
The end of the output should contain the counts of true positives, false positives, and false negatives. \\
The total count: \\
  - TP total count: \\
  - FP total count: \\
  - FN total count: \\
Ground truth: \{ground\_truth\} \\
Output: \{output\} \\

\textbf{\large{Model Response:}} \\
 1. **Identifying True Positives (TP)**\\
 - Heavy rainfall expected in the Mid-Atlantic aligns with both forecasts.\\
 - Heavy precipitation general trend noted in both forecasts.\\  - Snow and ice possibly in the Midwest and Northeast matches the mention of snow/ice in northern regions.\\
 2. **Identifying False Positives (FP)**\\
 - The predicted forecast assigns specific amounts of rainfall (2 to 4 inches), which is not indicated in the ground truth forecast, leading to revelation of specific metrics not originally stated.\\
 - The predicted forecast introduces a slight risk of excessive rainfall which is not mentioned at all in the ground truth forecast.\\
 3. **Identifying False Negatives (FN)**\\
 - The ground truth mentions the Ohio Valley and Central Appalachians but the predicted forecast shifts focus more towards southern Plains and Mid-South, dropping some regions. \\
 - Risks of isolated flash flooding due to saturated soils and specifics about heavy precipitation in areas like the Northern Rockies and Great Basin mentioned in the ground truth are neglected in the predicted forecast. \\
 - The reference to much below normal temperatures affecting parts of Alaska is not mirrored in the predicted forecast.\\
Summary Counts:\\
- TP total count: 3\\
- FP total count: 2\\
- FN total count: 3\\
\end{mdframed}
\newpage
\section{More Quantitative Results}\label{app:quant_results}
\subsection{Weather Dataset}
\begin{table}[htbp!]
  \caption{Cosine Similarity Scores for various models on climate dataset with different window sizes}
  \label{tab:climate-cos-sim-scores}
  \centering
  \small
  \vspace{0.5cm}
  \begin{tabular}{l|ccccccc}
    \toprule
    \textbf{Model} & \textbf{1-1} & \textbf{2-2} & \textbf{3-3} & \textbf{4-4} & \textbf{5-5} & \textbf{6-6} & \textbf{7-7} \\
    \midrule
    input\_copy & \textbf{0.824} & \underline{0.790} & 0.769 & 0.762 & 0.756 & 0.751 & 0.746 \\
    Text2Text(zero-shot) & 0.727 & 0.715 & 0.700 & 0.701 & 0.700 & 0.691 & 0.701 \\
    TextTime2Text (zero-shot) & 0.729 & 0.723 & 0.705 & 0.695 & 0.702 & 0.700 & 0.706 \\
    TextTime2TextTime (zero-shot) & 0.708 & 0.708 & 0.683 & 0.680 & 0.680 & 0.681 & 0.692 \\
    Text2Text(in-context) & 0.774 & 0.772 & 0.759 & 0.760 & 0.760 & 0.758 & 0.757 \\
    TextTime2Text (in-context) & 0.769 & 0.767 & 0.759 & 0.760 & 0.759 & 0.756 & 0.756 \\
    TextTime2TextTime (in-context) & 0.785 & 0.775 & 0.765 & 0.764 & 0.763 & 0.760 & 0.761 \\
    Text2Text (fine-tuned) & 0.791 & 0.784 & 0.779 & \underline{0.780} & 0.772 & 0.771 & \underline{0.771} \\
    TextTime2Text (fine-tuned) & 0.796 & 0.783 & \underline{0.783} & 0.779 & \underline{0.777} & 0.773 & \underline{0.771} \\
    TextTime2TextTime (fine-tuned) & 0.793 & 0.777 & 0.779 & 0.778 & 0.773 & \underline{0.774} & \underline{0.771} \\
    Hybrid & \underline{0.806} & \textbf{0.803} & \textbf{0.787} & \textbf{0.797} & \textbf{0.786} & \textbf{0.780} & \textbf{0.780} \\
    \bottomrule
  \end{tabular}
\end{table}
\begin{table}[htbp!]
  \caption{GPT Score (output single value) for various models on climate dataset with different window sizes}
  \label{tab:climate-gpt-score}
  \centering
  \small
  \vspace{0.5cm}
  \begin{tabular}{l|ccccccc}
    \toprule
    \textbf{Model} & \textbf{1-1} & \textbf{2-2} & \textbf{3-3} & \textbf{4-4} & \textbf{5-5} & \textbf{6-6} & \textbf{7-7} \\
    \midrule
        input\_copy & \textbf{7.683} & 6.584 & 5.906 & 5.308 & 4.801 & 4.401 & 4.220 \\
        Text2Text(zero-shot) & 4.678 & 4.618 & 4.658 & 4.478 & 4.365 & 4.430 & 4.575 \\
        TextTime2Text(zero-shot) & 4.994 & 5.235 & 4.655 & 4.315 & 4.476 & 4.596 & 4.472 \\
        TextTime2TextTime (zero-shot) & 4.227 & 4.650 & 4.170 & 3.958 & 3.920 & 3.808 & 3.951 \\
        TextText(in-context) & 7.380 & 6.482 & 6.511 & 6.131 & 6.029 & 5.913 & 5.721 \\
        TextTime2Text(in-context) & 7.284 & 6.441 & 6.247 & 5.952 & 5.835 & 5.573 & 5.523 \\
        TextTime2TextTime (in-context) & 6.771 & 6.351 & 6.029 & 5.801 & 5.652 & 5.563 & 5.616 \\
        Text2Text(fine-tuned) & 7.336 & \underline{7.108} & \underline{6.866} & \underline{6.740} & 6.583 & \textbf{6.235} & 6.030 \\
        TextTime2Text(fine-tuned) & 7.248 & \textbf{7.133} & 6.789 & \textbf{6.742} & \textbf{6.676} & 6.198 & \underline{6.337} \\
        TextTime2TextTime (fine-tuned) & \underline{7.424} & 7.086 & \textbf{6.891} & 6.738 & \underline{6.585} & \underline{6.204} & \textbf{6.365} \\
        Hybrid & 5.096 & 5.126 & 5.152 & 5.394 & 5.867 & 6.025 & 6.235 \\
    \bottomrule
  \end{tabular}
\end{table}
\begin{table}[htbp!]
  \caption{GPT Score Precision for various models on climate dataset with different window sizes}
  \label{tab:climate-gpt-score-precision}
  \centering
  \small
  \vspace{0.5cm}
  \begin{tabular}{l|ccccccc}
    \toprule
    \textbf{Model} & \textbf{1-1} & \textbf{2-2} & \textbf{3-3} & \textbf{4-4} & \textbf{5-5} & \textbf{6-6} & \textbf{7-7} \\
    \midrule
    input\_copy & \textbf{0.754} & \textbf{0.696} & 0.638 & 0.613 & 0.588 & 0.568 & 0.566 \\
    Text2Text(zero-shot) & 0.707 & 0.636 & 0.606 & 0.552 & 0.548 & 0.512 & 0.516 \\
    TextTime2Text (zero-shot) & 0.663 & 0.665 & 0.598 & 0.553 & 0.550 & 0.497 & 0.539 \\
    TextTime2TextTime (zero-shot) & 0.710 & 0.640 & 0.580 & 0.560 & 0.550 & 0.469 & 0.458 \\
    Text2Text(in-context) & 0.708 & 0.638 & 0.640 & 0.602 & 0.603 & 0.590 & 0.574 \\
    TextTime2Text (in-context) & 0.701 & 0.646 & 0.621 & 0.605 & 0.596 & 0.580 & 0.575 \\
    TextTime2TextTime (in-context) & \underline{0.727} & 0.666 & 0.641 & 0.619 & 0.598 & 0.580 & 0.578 \\
    Text2Text (fine-tuned) & 0.695 & 0.684 & 0.662 & \underline{0.647} & \textbf{0.639} & 0.609 & 0.589 \\
    TextTime2Text (fine-tuned) & 0.678 & 0.682 & 0.655 & 0.639 & \underline{0.629} & 0.598 & \underline{0.608} \\
    TextTime2TextTime (fine-tuned) & 0.710 & \underline{0.687} & \underline{0.668} & \textbf{0.650} & \underline{0.629} & \underline{0.614} & \textbf{0.609} \\
    Hybrid & 0.709 & 0.678 & \textbf{0.677} & 0.638 & 0.618 & \textbf{0.615} & 0.606 \\
    \bottomrule
  \end{tabular}
\end{table}
\begin{table}[htbp!]
  \caption{GPT F1 Recall for various models on climate dataset with different window sizes}
  \label{tab:climate-gpt-f1-recall}
  \centering
  \small
  \vspace{0.5cm}
  \begin{tabular}{l|ccccccc}
    \toprule
    \textbf{Model} & \textbf{1-1} & \textbf{2-2} & \textbf{3-3} & \textbf{4-4} & \textbf{5-5} & \textbf{6-6} & \textbf{7-7} \\
    \midrule
    input\_copy & \textbf{0.599} & 0.522 & 0.462 & 0.433 & 0.407 & 0.389 & 0.391 \\
    Text2Text(zero-shot) & 0.333 & 0.314 & 0.306 & 0.287 & 0.275 & 0.275 & 0.297 \\
    TextTime2Text (zero-shot) & 0.332 & 0.359 & 0.305 & 0.275 & 0.281 & 0.286 & 0.286 \\
    TextTime2TextTime (zero-shot) & 0.292 & 0.300 & 0.257 & 0.245 & 0.239 & 0.217 & 0.223 \\
    Text2Text(in-context) & 0.512 & 0.450 & 0.450 & 0.415 & 0.416 & 0.400 & 0.392 \\
    TextTime2Text (in-context) & 0.503 & 0.444 & 0.431 & 0.412 & 0.403 & 0.382 & 0.384 \\
    TextTime2TextTime (in-context) & 0.486 & 0.452 & 0.430 & 0.409 & 0.395 & 0.384 & 0.381 \\
    Text2Text (fine-tuned) & 0.558 & 0.522 & 0.503 & \underline{0.496} & 0.475 & 0.460 & 0.438 \\
    TextTime2Text (fine-tuned) & 0.547 & \underline{0.528} & 0.505 & 0.484 & \underline{0.477} & 0.458 & \underline{0.457} \\
    TextTime2TextTime (fine-tuned) & \underline{0.569} & 0.527 & \underline{0.508} & 0.494 & \underline{0.477} & \underline{0.462} & 0.455 \\
    Hybrid & 0.557 & \textbf{0.539} & \textbf{0.536} & \textbf{0.502} & \textbf{0.484} & \textbf{0.472} & \textbf{0.468} \\
    \bottomrule
  \end{tabular}
\end{table}
\begin{table}[htbp!]
  \caption{GPT Score Recall for various models on climate dataset with different window sizes}
  \label{tab:climate-gpt-score-recall}
  \centering
  \small
  \vspace{0.5cm}
  \begin{tabular}{l|ccccccc}
    \toprule
    \textbf{Model} & \textbf{1-1} & \textbf{2-2} & \textbf{3-3} & \textbf{4-4} & \textbf{5-5} & \textbf{6-6} & \textbf{7-7} \\
    \midrule
    input\_copy & \textbf{0.521} & 0.446 & 0.393 & 0.365 & 0.343 & 0.326 & 0.329 \\
    Text2Text(zero-shot) & 0.241 & 0.231 & 0.226 & 0.212 & 0.200 & 0.209 & 0.231 \\
    TextTime2Text (zero-shot) & 0.245 & 0.269 & 0.226 & 0.201 & 0.209 & 0.222 & 0.215 \\
    TextTime2TextTime (zero-shot) & 0.201 & 0.214 & 0.180 & 0.174 & 0.168 & 0.156 & 0.164 \\
    Text2Text(in-context) & 0.431 & 0.375 & 0.373 & 0.343 & 0.346 & 0.329 & 0.325 \\
    TextTime2Text (in-context) & 0.420 & 0.364 & 0.358 & 0.341 & 0.333 & 0.313 & 0.316 \\
    TextTime2TextTime (in-context) & 0.397 & 0.367 & 0.352 & 0.334 & 0.322 & 0.314 & 0.310 \\
    Text2Text (fine-tuned) & 0.497 & 0.455 & 0.438 & \underline{0.434} & 0.413 & 0.402 & 0.380 \\
    TextTime2Text (fine-tuned) & 0.485 & \underline{0.462} & \underline{0.443} & 0.421 & 0.414 & \underline{0.403} & \underline{0.398} \\
    TextTime2TextTime (fine-tuned) & \underline{0.507} & 0.459 & 0.442 & 0.430 & \underline{0.416} & \underline{0.403} & 0.394 \\
    Hybrid & 0.494 & \textbf{0.475} & \textbf{0.474} & \textbf{0.447} & \textbf{0.428} & \textbf{0.417} & \textbf{0.417} \\
    \bottomrule
  \end{tabular}
\end{table}
\begin{table}[htbp!]
  \caption{Meteor Scores for various models on climate dataset with different window sizes. This is a text evaluation metric with Meteor,
  }
  \label{tab:climate-meteor-scores}
  \centering
  \small
  \vspace{0.5cm}
  \begin{tabular}{l|ccccccc}
    \toprule
    \textbf{Model} & \textbf{1-1} & \textbf{2-2} & \textbf{3-3} & \textbf{4-4} & \textbf{5-5} & \textbf{6-6} & \textbf{7-7} \\
    \midrule
    input\_copy & \textbf{0.397} & 0.359 & 0.341 & 0.334 & 0.329 & 0.328 & 0.330 \\
    Text2Text(zero-shot) & 0.174 & 0.180 & 0.172 & 0.165 & 0.161 & 0.155 & 0.172 \\
    TextTime2Text (zero-shot) & 0.187 & 0.198 & 0.180 & 0.163 & 0.173 & 0.170 & 0.183 \\
    TextTime2TextTime (zero-shot) & 0.133 & 0.160 & 0.145 & 0.136 & 0.140 & 0.140 & 0.159 \\
    Text2Text(in-context) & 0.281 & 0.273 & 0.265 & 0.268 & 0.265 & 0.257 & 0.273 \\
    TextTime2Text (in-context) & 0.269 & 0.263 & 0.260 & 0.266 & 0.263 & 0.256 & 0.269 \\
    TextTime2TextTime (in-context) & 0.287 & 0.281 & 0.277 & 0.274 & 0.269 & 0.264 & 0.272 \\
    Text2Text (fine-tuned) & 0.374 & \textbf{0.370} & \underline{0.368} & 0.363 & 0.362 & 0.359 & \textbf{0.359} \\
    TextTime2Text (fine-tuned) & 0.374 & \textbf{0.370} & \textbf{0.369} & \textbf{0.365} & \textbf{0.364} & \underline{0.361} & \underline{0.358} \\
    TextTime2TextTime (fine-tuned) & 0.375 & \underline{0.364} & \underline{0.368} & \underline{0.364} & \underline{0.363} & \textbf{0.362} & \textbf{0.359} \\
    Hybrid & \underline{0.383} & 0.359 & 0.363 & 0.348 & 0.338 & 0.336 & 0.348 \\
    \bottomrule
  \end{tabular}
\end{table}
\begin{table}[htbp!]
  \caption{Rouge1 scores for various models on climate dataset with different window sizes}
  \label{tab:climate-rouge1-scores}
  \centering
  \small
  \vspace{0.5cm}
  \begin{tabular}{l|ccccccc}
    \toprule
    \textbf{Model} & \textbf{1-1} & \textbf{2-2} & \textbf{3-3} & \textbf{4-4} & \textbf{5-5} & \textbf{6-6} & \textbf{7-7} \\
    \midrule
    input\_copy & \textbf{0.572} & 0.533 & 0.516 & 0.503 & 0.499 & 0.497 & 0.496 \\
    Text2Text(zero-shot) & 0.326 & 0.340 & 0.326 & 0.319 & 0.314 & 0.304 & 0.333 \\
    TextTime2Text (zero-shot) & 0.343 & 0.363 & 0.334 & 0.310 & 0.326 & 0.322 & 0.341 \\
    TextTime2TextTime (zero-shot) & 0.268 & 0.306 & 0.279 & 0.267 & 0.273 & 0.272 & 0.301 \\
    Text2Text(in-context) & 0.494 & 0.479 & 0.475 & 0.470 & 0.468 & 0.459 & 0.466 \\
    TextTime2Text (in-context) & 0.482 & 0.469 & 0.466 & 0.467 & 0.460 & 0.450 & 0.459 \\
    TextTime2TextTime (in-context) & 0.481 & 0.478 & 0.471 & 0.467 & 0.461 & 0.456 & 0.463 \\
    Text2Text (fine-tuned) & 0.552 & \textbf{0.540} & \textbf{0.538} & \underline{0.532} & \underline{0.530} & \underline{0.530} & \textbf{0.529} \\
    TextTime2Text (fine-tuned) & 0.551 & \underline{0.538} & \underline{0.537} & \textbf{0.533} & \textbf{0.531} & \textbf{0.531} & \underline{0.527} \\
    TextTime2TextTime (fine-tuned) & 0.550 & 0.533 & 0.536 & \textbf{0.533} & 0.530 & 0.530 & 0.526 \\
    Hybrid & \underline{0.566} & 0.495 & 0.485 & 0.480 & 0.471 & 0.487 & 0.501 \\
    \bottomrule
  \end{tabular}
\end{table}
\begin{table}[htbp!]
  \caption{Rouge2 Scores for various models on climate dataset with different window sizes}
  \label{tab:climate-rouge2-scores}
  \centering
  \small
  \vspace{0.5cm}
  \begin{tabular}{l|ccccccc}
    \toprule
    \textbf{Model} & \textbf{1-1} & \textbf{2-2} & \textbf{3-3} & \textbf{4-4} & \textbf{5-5} & \textbf{6-6} & \textbf{7-7} \\
    \midrule
    input\_copy & \underline{0.242} & 0.197 & 0.175 & 0.164 & 0.158 & 0.157 & 0.155 \\
    Text2Text(zero-shot) & 0.157 & 0.148 & 0.136 & 0.135 & 0.132 & 0.127 & 0.134 \\
    TextTime2Text (zero-shot) & 0.163 & 0.157 & 0.140 & 0.131 & 0.136 & 0.133 & 0.136 \\
    TextTime2TextTime (zero-shot) & 0.135 & 0.138 & 0.123 & 0.119 & 0.119 & 0.119 & 0.125 \\
    Text2Text(in-context) & 0.202 & 0.186 & 0.177 & 0.175 & 0.174 & 0.170 & 0.171 \\
    TextTime2Text (in-context) & 0.199 & 0.184 & 0.177 & 0.176 & 0.173 & 0.170 & 0.171 \\
    TextTime2TextTime (in-context) & 0.206 & 0.194 & 0.182 & 0.179 & 0.177 & 0.173 & 0.172 \\
    Text2Text (fine-tuned) & 0.230 & \textbf{0.221} & \textbf{0.218} & \underline{0.212} & 0.209 & \textbf{0.208} & \underline{0.205} \\
    TextTime2Text (fine-tuned) & 0.231 & \underline{0.220} & \underline{0.217} & \textbf{0.213} & \textbf{0.211} & \textbf{0.208} & \underline{0.205} \\
    TextTime2TextTime (fine-tuned) & 0.230 & 0.217 & \underline{0.217} & \textbf{0.213} & \underline{0.210} & \textbf{0.208} & 0.204 \\
    Hybrid & \textbf{0.253} & 0.216 & 0.205 & 0.208 & 0.199 & \underline{0.202} & \textbf{0.206} \\
    \bottomrule
  \end{tabular}
\end{table}
\begin{table}[htbp!]
  \caption{RougeL Scores for various models on climate dataset with different window sizes}
  \label{tab:climate-rougel-scores}
  \centering
  \small
  \vspace{0.5cm}
  \begin{tabular}{l|ccccccc}
    \toprule
    \textbf{Model} & \textbf{1-1} & \textbf{2-2} & \textbf{3-3} & \textbf{4-4} & \textbf{5-5} & \textbf{6-6} & \textbf{7-7} \\
    \midrule
    input\_copy & \underline{0.297} & 0.260 & 0.241 & 0.234 & 0.229 & 0.227 & 0.226 \\
    Text2Text(zero-shot) & 0.217 & 0.214 & 0.203 & 0.202 & 0.200 & 0.195 & 0.204 \\
    TextTime2Text (zero-shot) & 0.222 & 0.222 & 0.205 & 0.198 & 0.203 & 0.201 & 0.207 \\
    TextTime2TextTime (zero-shot) & 0.196 & 0.202 & 0.186 & 0.182 & 0.184 & 0.183 & 0.192 \\
    Text2Text(in-context) & 0.269 & 0.258 & 0.250 & 0.248 & 0.249 & 0.245 & 0.243 \\
    TextTime2Text (in-context) & 0.267 & 0.255 & 0.250 & 0.249 & 0.247 & 0.244 & 0.244 \\
    TextTime2TextTime (in-context) & 0.269 & 0.260 & 0.250 & 0.249 & 0.248 & 0.246 & 0.244 \\
    Text2Text (fine-tuned) & 0.281 & \textbf{0.275} & \underline{0.271} & \underline{0.268} & \textbf{0.266} & \textbf{0.265} & \underline{0.262} \\
    TextTime2Text (fine-tuned) & 0.284 & \underline{0.274} & \textbf{0.272} & \underline{0.268} & \textbf{0.266} & \textbf{0.265} & \underline{0.262} \\
    TextTime2TextTime (fine-tuned) & 0.280 & 0.271 & \underline{0.271} & \textbf{0.269} & \textbf{0.266} & \textbf{0.265} & 0.261 \\
    Hybrid & \textbf{0.304} & 0.271 & 0.259 & 0.260 & \underline{0.255} & \underline{0.260} & \textbf{0.264} \\
    \bottomrule
  \end{tabular}
\end{table}

\newpage
\subsection{Medical Dataset}
\begin{table}[h!]
  \caption{Cosine similarity scores for various models on medical dataset with different window sizes}
  \label{tab:medical-cos-sim}
  \centering
  \small
  \vspace{0.5cm}
  \begin{tabular}{l|cccccc}
    \toprule
    \textbf{Model} & \textbf{1-1} & \textbf{2-2} & \textbf{3-3} & \textbf{4-4} & \textbf{5-5} & \textbf{6-6}\\
    \midrule
    input\_copy & 5.235 & 5.909 & 6.084 & 6.544 & 7.094 & 7.479 \\
    nlinear & 5.195 & 5.279 & 5.275 & 5.406 & \underline{5.562} & \underline{5.875} \\
    nlinear\_newer\_embedding(bge-small-en-v1.5) & \underline{5.117} & \underline{5.143} & \textbf{5.106} & \textbf{5.300} & \textbf{5.492} & \textbf{5.759} \\
    TextTime2TextTime (zero-shot) & 11.549 & 8.664 & 8.388 & 9.057 & 9.476 & 8.963 \\
    TextTime2Time (zero-shot) & 6.706 & 6.008 & 7.529 & 7.179 & 7.935 & 8.734 \\
    TextTime2TextTime (in-context) & 6.600 & 6.620 & 6.521 & 7.291 & 7.967 & 8.165 \\
    TextTime2Time (in-context) & 7.182 & 6.414 & 6.479 & 7.226 & 7.527 & 7.886 \\
    TextTime2TextTime (fine-tuned) & 6.689 & 6.432 & 6.022 & 6.483 & 6.747 & 7.631 \\
    TextTime2Time (fine-tuned) & 6.084 & 6.246 & 6.202 & 6.709 & 7.298 & 7.157 \\
    PatchTST & \textbf{4.659} & \textbf{4.807} & \underline{5.117} & \underline{5.317} & 5.462 & 5.798 \\
    Hybrid w/ llama tokens & 5.202 & 5.472 & 6.620 & 6.269 & 6.127 & 7.960 \\
    Hybrid w/ bge embeddings & 5.244 & 5.328 & 5.776 & 6.269 & 6.096 & 6.797 \\
    Hybrid w/ bge + w/2 & 5.238 & 5.313 & 5.782 & 5.647 & 6.121 & 8.350 \\
    \bottomrule
  \end{tabular}
\end{table}
\begin{table}[h!]
  \caption{GPT Score for various models on medical dataset with different window sizes}
  \label{tab:medical-gpt-score-single-value}
  \centering
  \small
  \vspace{0.5cm}
  \begin{tabular}{l|cccccc}
    \toprule
    \textbf{Model} & \textbf{1-1} & \textbf{2-2} & \textbf{3-3} & \textbf{4-4} & \textbf{5-5} & \textbf{6-6} \\
    \midrule
    input\_copy & 3.617 & 3.258 & 3.094 & 3.031 & 2.961 & 2.941 \\
    Text2Text(zero-shot) & 3.575 & 3.557 & 3.435 & 3.432 & 3.551 & 3.632 \\
    TextTime2Text (zero-shot) & 3.745 & 3.908 & 3.674 & 3.619 & 3.634 & 3.723 \\
    TextTime2TextTime (zero-shot) & \underline{3.969} & \underline{3.999} & \underline{3.852} & \underline{3.794} & \underline{3.803} & \textbf{4.018} \\
    Text2Text(in-context) & 3.352 & 3.288 & 3.394 & 3.510 & 3.534 & 3.557 \\
    TextTime2Text (in-context) & 3.424 & 3.351 & 3.435 & 3.492 & 3.532 & 3.544 \\
    TextTime2TextTime (in-context) & 3.432 & 3.385 & 3.393 & 3.424 & 3.439 & 3.476 \\
    Text2Text (fine-tuned) & 3.353 & 3.302 & 3.632 & 3.426 & 3.365 & 3.549 \\
    TextTime2Text (fine-tuned) & 3.308 & 3.359 & 3.396 & 3.411 & 3.397 & 3.536 \\
    TextTime2TextTime (fine-tuned) & 3.358 & 3.345 & 3.601 & 3.544 & 3.480 & 3.426 \\
    Hybrid w/ llama tokens & \textbf{4.215} & \textbf{4.117} & \textbf{4.102} & \textbf{4.048} & \textbf{4.043} & \underline{3.990} \\
    \bottomrule
  \end{tabular}
\end{table}
\begin{table}[h!]
  \caption{GPT Score Precision for various models on medical dataset with different window sizes}
  \label{tab:medical-gpt-score-precision}
  \centering
  \small
  \vspace{0.5cm}
  \begin{tabular}{l|cccccc}
    \toprule
    \textbf{Model} & \textbf{1-1} & \textbf{2-2} & \textbf{3-3} & \textbf{4-4} & \textbf{5-5} & \textbf{6-6} \\
    \midrule
    input\_copy & \textbf{0.548} & \textbf{0.524} & 0.520 & 0.501 & 0.499 & 0.488 \\
    Text2Text(zero-shot) & 0.493 & 0.451 & 0.418 & 0.425 & 0.426 & 0.422 \\
    TextTime2Text (zero-shot) & 0.493 & 0.492 & 0.461 & 0.440 & 0.436 & 0.440 \\
    TextTime2TextTime (zero-shot) & 0.499 & 0.502 & 0.470 & 0.458 & 0.470 & 0.484 \\
    Text2Text(in-context) & 0.453 & 0.442 & 0.441 & 0.449 & 0.452 & 0.452 \\
    TextTime2Text (in-context) & 0.465 & 0.451 & 0.445 & 0.448 & 0.466 & 0.456 \\
    TextTime2TextTime (in-context) & 0.457 & 0.467 & 0.469 & 0.467 & 0.455 & 0.467 \\
    Text2Text (fine-tuned) & \underline{0.519} & 0.506 & 0.392 & 0.508 & 0.516 & 0.507 \\
    TextTime2Text (fine-tuned) & 0.515 & \underline{0.515} & \textbf{0.531} & \textbf{0.528} & \underline{0.519} & \textbf{0.514} \\
    TextTime2TextTime (fine-tuned) & 0.518 & \textbf{0.524} & \underline{0.523} & \underline{0.518} & \textbf{0.520} & \textbf{0.514} \\
    Hybrid w/ llama tokens & 0.508 & 0.506 & 0.519 & 0.498 & 0.506 & 0.489 \\
    Hybrid w/ bge embeddings & 0.503 & 0.511 & 0.511 & 0.512 & 0.517 & 0.486 \\
    Hybrid w/ bge + weight = 2 & 0.516 & 0.508 & 0.521 & 0.517 & 0.499 & \underline{0.512} \\
    \bottomrule
  \end{tabular}
\end{table}
\begin{table}[h!]
  \caption{GPT F1 Recall for various models on medical dataset with different window sizes}
  \label{tab:medical-gpt-f1-recall}
  \centering
  \small
  \vspace{0.5cm}
  \begin{tabular}{l|ccccccc}
    \toprule
    \textbf{Model} & \textbf{1-1} & \textbf{2-2} & \textbf{3-3} & \textbf{4-4} & \textbf{5-5} & \textbf{6-6} & \textbf{7-7} \\
    \midrule
    input\_copy & 0.459 & 0.444 & 0.437 & 0.423 & 0.418 & 0.409 \\
    Text2Text(zero-shot) & 0.425 & 0.380 & 0.357 & 0.354 & 0.350 & 0.346 \\
    TextTime2Text (zero-shot) & 0.426 & 0.399 & 0.370 & 0.357 & 0.347 & 0.344 \\
    TextTime2TextTime (zero-shot) & 0.397 & 0.381 & 0.351 & 0.349 & 0.364 & 0.334 \\
    Text2Text(in-context) & 0.389 & 0.382 & 0.383 & 0.391 & 0.395 & 0.392 \\
    TextTime2Text (in-context) & 0.404 & 0.391 & 0.386 & 0.389 & 0.404 & 0.398 \\
    TextTime2TextTime (in-context) & 0.397 & 0.403 & 0.403 & 0.403 & 0.395 & 0.402 \\
    Text2Text (fine-tuned) & \textbf{0.466} & 0.463 & 0.357 & 0.457 & \textbf{0.470} & 0.461 \\
    TextTime2Text (fine-tuned) & \underline{0.461} & \underline{0.465} & \textbf{0.477} & \textbf{0.473} & 0.468 & \textbf{0.468} \\
    TextTime2TextTime (fine-tuned) & 0.458 & \textbf{0.474} & \underline{0.474} & \underline{0.469} & \underline{0.469} & \underline{0.466} \\
    Hybrid w/ llama tokens & 0.456 & 0.463 & 0.461 & 0.448 & 0.462 & 0.443 \\
    Hybrid w/ bge embeddings & 0.450 & 0.465 & 0.454 & 0.457 & 0.453 & 0.441 \\
    Hybrid w/ bge + weight = 2 & 0.460 & 0.447 & 0.471 & 0.459 & 0.452 & 0.459 \\
    \bottomrule
  \end{tabular}
\end{table}
\begin{table}[h!]
  \caption{GPT Score Recall for various models on medical dataset with different window sizes}
  \label{tab:medical-gpt-score-recall}
  \centering
  \small
  \vspace{0.5cm}
  \begin{tabular}{l|cccccc}
    \toprule
    \textbf{Model} & \textbf{1-1} & \textbf{2-2} & \textbf{3-3} & \textbf{4-4} & \textbf{5-5} & \textbf{6-6}\\
    \midrule
    input\_copy & 0.424 & 0.413 & 0.405 & 0.393 & 0.387 & 0.378 \\
    Text2Text(zero-shot) & 0.402 & 0.357 & 0.340 & 0.336 & 0.326 & 0.323 \\
    TextTime2Text (zero-shot) & 0.402 & 0.368 & 0.344 & 0.330 & 0.320 & 0.314 \\
    TextTime2TextTime (zero-shot) & 0.359 & 0.341 & 0.314 & 0.316 & 0.333 & 0.290 \\
    Text2Text(in-context) & 0.369 & 0.363 & 0.365 & 0.371 & 0.376 & 0.371 \\
    TextTime2Text (in-context) & 0.383 & 0.371 & 0.368 & 0.371 & 0.384 & 0.377 \\
    TextTime2TextTime (in-context) & 0.375 & 0.381 & 0.380 & 0.379 & 0.374 & 0.378 \\
    Text2Text (fine-tuned) & \textbf{0.447} & 0.450 & 0.347 & 0.442 & \textbf{0.457} & 0.448 \\
    TextTime2Text (fine-tuned) & 0.441 & 0.447 & \textbf{0.460} & \underline{0.454} & 0.451 & \textbf{0.455} \\
    TextTime2TextTime (fine-tuned) & 0.435 & \textbf{0.460} & \underline{0.456} & \textbf{0.455} & \underline{0.452} & \underline{0.453} \\
    Hybrid w/ llama tokens & 0.437 & 0.453 & 0.440 & 0.438 & 0.448 & 0.428 \\
    Hybrid w/ bge embeddings & 0.433 & \underline{0.454} & 0.434 & 0.439 & 0.448 & 0.428 \\
    Hybrid w/ bge + weight = 2 & \underline{0.442} & 0.424 & 0.454 & 0.438 & 0.437 & 0.448 \\
    \bottomrule
  \end{tabular}
\end{table}
\begin{table}[h!]
  \caption{Meteor Scores for various models on medical dataset with different window sizes}
  \label{tab:medical-meteor-scores}
  \centering
  \small
  \vspace{0.5cm}
  \begin{tabular}{l|ccccccc}
    \toprule
    \textbf{Model} & \textbf{1-1} & \textbf{2-2} & \textbf{3-3} & \textbf{4-4} & \textbf{5-5} & \textbf{6-6} \\
    \midrule
    input\_copy & 0.422 & 0.418 & 0.412 & 0.413 & 0.412 & 0.411 \\
    TextTime2TextTime (zero-shot) & 0.416 & 0.374 & 0.360 & 0.344 & 0.326 & 0.306 \\
    TextTime2TextTime (zero-shot) & 0.405 & 0.332 & 0.329 & 0.323 & 0.303 & 0.285 \\
    TextTime2TextTime (zero-shot) & 0.335 & 0.298 & 0.273 & 0.286 & 0.297 & 0.230 \\
    TextTime2TextTime (in-context) & 0.408 & 0.401 & 0.386 & 0.375 & 0.377 & 0.373 \\
    TextTime2TextTime (in-context) & 0.412 & 0.402 & 0.390 & 0.380 & 0.382 & 0.384 \\
    TextTime2TextTime (fine-tuned) & \textbf{0.430} & 0.441 & 0.414 & 0.439 & \textbf{0.440} & 0.420 \\
    TextTime2TextTime (fine-tuned) & 0.426 & 0.443 & \underline{0.439} & 0.437 & \underline{0.439} & 0.435 \\
    TextTime2TextTime (fine-tuned) & 0.427 & 0.440 & 0.405 & 0.422 & 0.429 & 0.429 \\
    Hybrid w/ llama tokens & 0.427 & 0.426 & \textbf{0.452} & \underline{0.446} & \underline{0.439} & \textbf{0.443} \\
    Hybrid w/ bge embeddings & \underline{0.429} & \underline{0.445} & 0.432 & \textbf{0.450} & 0.428 & 0.438 \\
    Hybrid w/ bge + weight = 2 & 0.428 & \textbf{0.446} & 0.435 & 0.416 & 0.438 & \underline{0.440} \\
    \bottomrule
  \end{tabular}
\end{table}
\begin{table}[h!]
  \caption{Rouge1 Scores for various models on medical dataset with different window sizes}
  \label{tab:medical-rouge1-scores}
  \centering
  \small
  \vspace{0.5cm}
  \begin{tabular}{l|ccccccc}
    \toprule
    \textbf{Model} & \textbf{1-1} & \textbf{2-2} & \textbf{3-3} & \textbf{4-4} & \textbf{5-5} & \textbf{6-6} \\
    \midrule
    input\_copy & 0.433 & 0.427 & 0.424 & 0.423 & 0.420 & 0.417 \\
    TextTime2Text(zero-shot) & 0.419 & 0.405 & 0.397 & 0.390 & 0.383 & 0.377 \\
    TextTime2Text (zero-shot) & 0.413 & 0.390 & 0.386 & 0.384 & 0.378 & 0.370 \\
    TextTime2TextTime (zero-shot) & 0.385 & 0.375 & 0.362 & 0.370 & 0.378 & 0.345 \\
    Text2Text(in-context) & 0.419 & 0.420 & 0.417 & 0.411 & 0.414 & 0.411 \\
    TextTime2Text (in-context) & 0.420 & 0.423 & 0.420 & 0.416 & 0.417 & 0.417 \\
    TextTime2TextTime (in-context) & 0.421 & 0.425 & 0.426 & 0.428 & 0.428 & 0.429 \\
    Text2Text (fine-tuned) & 0.449 & 0.449 & 0.433 & 0.449 & 0.447 & 0.438 \\
    TextTime2Text (fine-tuned) & 0.446 & 0.451 & 0.450 & 0.448 & 0.446 & 0.448 \\
    TextTime2TextTime (fine-tuned) & 0.448 & 0.449 & 0.430 & 0.437 & 0.441 & 0.444 \\
    Hybrid w/ llama tokens & \textbf{0.466} & \textbf{0.464} & \textbf{0.465} & \underline{0.462} & \underline{0.459} & 0.450 \\
    Hybrid w/ bge embeddings & 0.463 & 0.454 & 0.461 & \underline{0.462} & 0.457 & \textbf{0.458} \\
    Hybrid w/ bge + weight = 2 & \underline{0.464} & \underline{0.456} & \underline{0.461} & \textbf{0.465} & \textbf{0.460} & \underline{0.456} \\
    \bottomrule
  \end{tabular}
\end{table}
\begin{table}[h!]
  \caption{Rouge2 Scores for various models on medical dataset with different window sizes}
  \label{tab:medical-rouge2-scores}
  \centering
  \small
  \vspace{0.5cm}
  \begin{tabular}{l|ccccccc}
    \toprule
    \textbf{Model} & \textbf{1-1} & \textbf{2-2} & \textbf{3-3} & \textbf{4-4} & \textbf{5-5} & \textbf{6-6} \\
    \midrule
    input\_copy & 0.148 & 0.141 & 0.137 & 0.135 & 0.134 & 0.133 \\
    TextTime2Text(zero-shot) & 0.141 & 0.139 & 0.134 & 0.132 & 0.127 & 0.125 \\
    TextTime2Text (zero-shot) & 0.138 & 0.136 & 0.134 & 0.133 & 0.131 & 0.130 \\
    TextTime2TextTime (zero-shot) & 0.125 & 0.133 & 0.128 & 0.131 & 0.135 & 0.124 \\
    Text2Text(in-context) & 0.145 & 0.147 & 0.144 & 0.141 & 0.143 & 0.142 \\
    TextTime2Text (in-context) & 0.146 & 0.149 & 0.148 & 0.146 & 0.147 & 0.149 \\
    TextTime2TextTime (in-context) & 0.148 & 0.150 & 0.152 & 0.155 & 0.155 & 0.158 \\
    Text2Text (fine-tuned) & 0.179 & 0.177 & 0.164 & 0.176 & 0.175 & 0.168 \\
    TextTime2Text (fine-tuned) & 0.181 & 0.179 & 0.179 & 0.177 & 0.175 & 0.175 \\
    TextTime2TextTime (fine-tuned) & 0.181 & 0.179 & 0.163 & 0.168 & 0.172 & 0.172 \\
    Hybrid w/ llama tokens & \textbf{0.210} & \textbf{0.214} & 0.202 & \underline{0.203} & 0.203 & 0.197 \\
    Hybrid w/ bge embeddings & \underline{0.209} & 0.201 & \textbf{0.205} & \underline{0.203} & \textbf{0.209} & \textbf{0.207} \\
    Hybrid w/ bge + weight = 2 & \underline{0.209} & \underline{0.202} & \underline{0.204} & \textbf{0.215} & \underline{0.207} & \underline{0.203} \\
    \bottomrule
  \end{tabular}
\end{table}
\begin{table}[h!]
  \caption{RougeL Scores for various models on medical dataset with different window sizes}
  \label{tab:medical-rougel-scores}
  \centering
  \small
  \vspace{0.5cm}
  \begin{tabular}{l|ccccccc}
    \toprule
    \textbf{Model} & \textbf{1-1} & \textbf{2-2} & \textbf{3-3} & \textbf{4-4} & \textbf{5-5} & \textbf{6-6}\\
    \midrule
    input\_copy & 0.261 & 0.258 & 0.253 & 0.252 & 0.251 & 0.248 \\
    Text2Text(zero-shot) & 0.253 & 0.248 & 0.244 & 0.241 & 0.238 & 0.234 \\
    TextTime2Text (zero-shot) & 0.248 & 0.243 & 0.241 & 0.241 & 0.238 & 0.236 \\
    TextTime2TextTime (zero-shot) & 0.233 & 0.236 & 0.231 & 0.235 & 0.240 & 0.224 \\
    Text2Text(in-context) & 0.255 & 0.259 & 0.257 & 0.254 & 0.256 & 0.255 \\
    TextTime2Text (in-context) & 0.255 & 0.261 & 0.260 & 0.259 & 0.259 & 0.261 \\
    TextTime2TextTime (in-context) & 0.258 & 0.261 & 0.264 & 0.267 & 0.268 & 0.271 \\
    Text2Text (fine-tuned) & 0.294 & 0.290 & 0.275 & 0.287 & 0.286 & 0.280 \\
    TextTime2Text (fine-tuned) & 0.293 & 0.291 & 0.289 & 0.288 & 0.287 & 0.287 \\
    TextTime2TextTime (fine-tuned) & 0.292 & 0.289 & 0.275 & 0.279 & 0.282 & 0.266 \\
    Hybrid w/ llama tokens & 0.316 & \underline{0.319} & 0.309 & 0.305 & 0.309 & 0.302 \\
    Hybrid w/ bge embeddings & \textbf{0.318} & 0.307 & \textbf{0.314} & \underline{0.306} & \textbf{0.314} & \textbf{0.311} \\
    Hybrid w/ bge + weight = 2 & \underline{0.317} & \underline{0.309} & \underline{0.312} & \textbf{0.319} & \underline{0.312} & \underline{0.309} \\
    \bottomrule
  \end{tabular}
\end{table}

\end{document}